\documentclass[pdflatex,sn-mathphys-num]{sn-jnl}

\usepackage{graphicx}%
\usepackage{multirow}%
\usepackage{amsmath,amssymb,amsfonts}%
\usepackage{amsthm}%
\usepackage{mathrsfs}%
\usepackage[title]{appendix}%
\usepackage{xcolor}%
\usepackage{textcomp}%
\usepackage{manyfoot}%
\usepackage{booktabs}%
\usepackage{algorithm}%
\usepackage{algorithmicx}%
\usepackage[noend]{algpseudocode}%
\usepackage{listings}%
\usepackage{xspace}
\usepackage{bm}
\usepackage{makecell}
\usepackage{lmodern}

\usepackage{hyperref}
\let\svthefootnote\thefootnote
\newcommand\freefootnote[1]{%
  \let\thefootnote\relax%
  \footnotetext{#1}%
  \let\thefootnote\svthefootnote%
}

\newcommand{\A}{\mathcal{A}}
\newcommand{\D}{\mathcal{D}}

\newtheorem{definition}{Definition}%
\newcommand{\myparagraph}[1]{\smallskip\noindent\textbf{#1.}\xspace}

\DeclareMathOperator*{\argmin}{arg\,min}

\begin{document}


\title[P²CE: Model-Agnostic Plausible Pareto-Optimal Counterfactual Explanations]{P²CE: Model-Agnostic Plausible Pareto-Optimal Counterfactual Explanations}


\author[1]{\fnm{Arthur} \sur{Hendricks Mendes de Oliveira}}\email{a217048@dac.unicamp.br}
\equalcont{These authors contributed equally to this work.}

\author*[1]{\fnm{Giovani} \sur{Valdrighi}}\email{giovani.valdrighi@ic.unicamp.br}
\equalcont{These authors contributed equally to this work.}

\author[1]{\fnm{Marcos} \sur{Medeiros Raimundo}}\email{mraimundo@ic.unicamp.br}

\affil[1]{\orgdiv{Instituto de Computação}, \orgname{Universidade Estadual de Campinas}, \orgaddress{\street{Albert Einstein Avenue}, \city{Campinas}, \postcode{13083-889}, \state{São Paulo}, \country{Brazil}}}




\abstract{
The increasing use of machine learning algorithms in social applications has raised concerns about fairness and transparency, leading to the development of counterfactual explanations. These explanations supports individuals to understand and potentially alter unfavorable decisions in areas such as loan applications, job selections, and more, by providing actionable changes to input features that would lead to a desired outcome. Existing methods often struggle to balance feasibility, plausibility, and computational efficiency. To address this, we introduce P²CE, an algorithm for generating plausible Pareto-optimal counterfactual explanations, offering users a diverse set of optimal trade-offs between different notions of feasibility. P²CE employs an auxiliary isolation forest outlier detector to ensure that explanations are in accordance with the data distribution and leverages SHAP values to obtain optimal results with short computing times, regardless of the underlying model.  Our algorithm was empirically evaluated on three datasets, demonstrating superior performance in terms of both solution quality and computational efficiency compared to related techniques.}

\keywords{Counterfactual Explanations, Multi-objective, Outlier Detection}



\maketitle

\freefootnote{This is a pre-peer review, pre-print version of this article.}

\newpage 
\section{Introduction}
\begin{figure}[b]
    \centering
    \includegraphics[width=\linewidth]{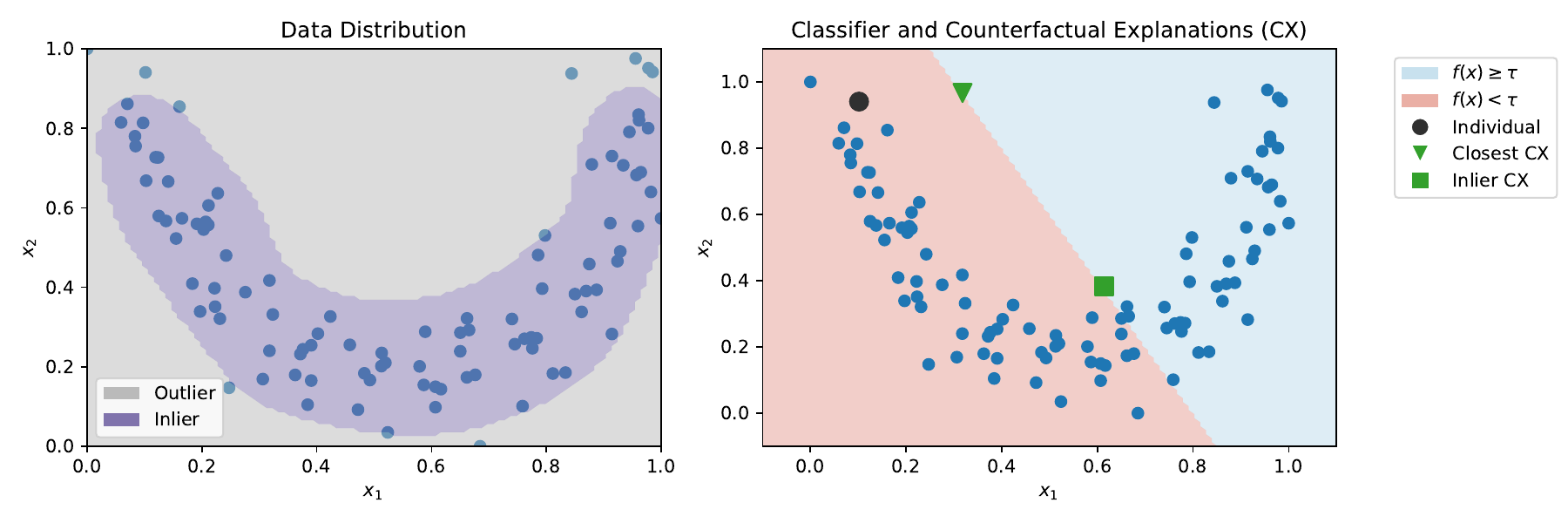}
    \caption{Toy example with a dataset of two features. On the left, we estimated the distribution using KDE. On the right, two algorithms generate one counterfactual explanation for the reference point in black. While the triangle explanation is close, it is outside the data distribution. The square point is the closest point inside the data distribution, it is feasible and plausible.}
    \label{fig:toy_example}
\end{figure}
\label{sec:introduction}

Machine learning algorithms are now present daily in the lives of individuals in many small and hidden decisions. However, these algorithms are also being used for decisions with a high impact on individuals, such as credit scoring, selection processes, health diagnosis, and others. In these contexts, supporting individuals in obtaining the desired prediction is important: ``My request was denied; what should I do next to have an accepted loan?''. Furthermore, regulations in diverse countries already ask for explanations on predictions, such as the EU General Data Protection Regulation (GDPR). Counterfactual explanations were proposed to answer such concerns~\cite{wachter2017counterfactual}, defined as changes in feature values necessary to obtain the desired prediction. The main objective of such explanations is to be truthful for the model's decision-making and to be \textbf{actionable}~\cite{karimi2021algorithmic}. Counterfactual explanations should be \textbf{feasible} ---suggested changes are achievable by the particular individual --- and should be \textbf{plausible} --- they are probable observations within the data distribution.

Given an observation $x$, counterfactual explanations can be expressed as a vector $x+a$ where $a$ represents the changes suggested for each feature. Previous works have been developed in this context, and different methods for measuring the feasibility of suggested changes $a$ have been discussed~\cite{verma2024counterfactual}. First, suggested changes should be small, as larger changes in features cannot be reachable for an individual (suggesting a ten-fold increase in income is not a useful counterfactual explanation, for example). Secondly, they should be sparse, suggesting changes in only a subset of features. This is motivated because a larger number of small changes may be worse compared to a moderate change on a smaller number of features. Furthermore, due to the subjective preference of individuals, no single measure of feasibility can be used for all scenarios. By providing multiple diverse solutions, different personal preferences can be supported. Previous works considered working with multiple objectives by defining a weighted loss~\cite{mothilal2020explaining} or employing multi-objective optimization to obtain a Pareto-optimal set of solutions~\cite{raimundo2022mining}.

However, when maximizing feasibility, algorithms might present solutions that are not likely to be based on the data distribution. In Fig.~\ref{fig:toy_example}, we present an example of counterfactual explanations with two features and Euclidean distance as a measure of feasibility. In this example, while the triangular solution might be closer in Euclidean distance to the reference sample (black dot), this solution is not inside the data distribution. The square solution, on the other hand, might be more distant from the reference sample; however, it ensures plausibility. In this simple example, the triangle solution was obtained by MAPOCAM~\cite{raimundo2022mining}, while the triangle algorithm is our solution. Previous algorithms have already been presented to ensure the plausibility of counterfactual explanations~\cite{kanamori2021dace, poyiadzi2020face}. However, they did not consider the importance of optimality regarding the multiple objectives.

Counterfactual explanation algorithms are also heavily dependent on the model considered. Depending on the complexity of the model, optimal counterfactual explanations might be challenging to obtain. Initial works presented solutions for differentiable and linear models, with more recent approaches being model-agnostic, considering that the model is a black box where only the predictions are known. Although these proposals may not achieve optimal efficiency across all models in computing cost, they are easily integrated into real-world applications.

We present P²CE, an algorithm designed to generate \textbf{P}lausible \textbf{P}areto-optimal \textbf{C}ounterfactual \textbf{E}xplanations. P²CE draws upon the previous work MAPOCAM but further improves the algorithm to ensure the plausibility of counterfactual explanations. The algorithm also addresses the limitations of prior research by enabling the generation of counterfactual explanations applicable to any model. Our proposal is based on a branch-and-bound search in a grid of possible counterfactual explanations. Still, it leverages outlier detection and SHAP values to avoid searching regions without useful solutions. This procedure can be done considering multiple objectives which commonly arise in the context of counterfactual explanations. P²CE obtains Pareto-optimal solutions; thus, no other solution in the space of counterfactual explanations would have costs lower or equal in all of the objectives. In summary, our contributions are as follows:

\begin{itemize}
    \item P²CE: A novel algorithm model-agnostic for the generation of optimal counterfactual explanations under multi-objectives that ensures that explanations are within the data distribution.
    \item The exploration of an unexplored SHAP's property that allows us to foresee if a set of variables can achieve a counterfactual. Thus expanding the agnostic capability of counterfactual explanations.
    \item An extensive evaluation of the proposed algorithm with multiple datasets and classifiers, compared with related techniques. The evaluation considers multi-objective values, computing time, and outlierness of solutions. 
    \item An open-source implementation of P²CE: \url{github.com/hiaac-finance/p2ce}
\end{itemize}

\section{Background and Related Works}
\label{sec:background}

We now formalize the problem of generating counterfactual explanations and present relevant works on the topic. In this section and in the remainder of this paper, we consider that we have access to a dataset $D = (X, Y); X \in \mathbb{R}^{n\times d}; Y \in \{0, 1\}^n$ sampled from a distribution $P$. We note as ($x_i, y_i$) the $i$-th sample of $D$ and $x(i)$ as the $i$-th coordinate of vector $x$. We also have a fitted classifier $f : \mathbb{R}^d \to [0, 1]$ that approximates the probability $f(x) \approx  P(Y = 1 \mid X = x)$. The classifier is accompanied by a threshold $\tau \in [0, 1]$, and a positive decision $\hat y_i = 1$ is made when $f(x_i) \geq \tau$. 

\subsection{Explanations}

Explainability has emerged as a very relevant area of research in machine learning as complex models become widely employed in applications that greatly impact individuals' lives. ``Explanations'' are presented in different forms, but, in general, they try to answer the question ``which features of the input were determinants of the output of the model?''. These explanations can be ``global'', with a general description of the relevant features for any input~\cite{friedman2001greedy, goldstein2015peeking}, or ``local'' where for a specific observation $x$, it is desired to obtain the set of relevant features. LIME~\cite{ribeiro2016should} and SHAP~\cite{lundberg2017unified} are among the most widely used methods for local explanations and present a similar concept: studying how small perturbations in each feature impact the model prediction. For a given observation $x$ and a model $f$, a vector $\phi \in \mathbb{R}^d$ is returned where $\phi(i)$ represents the ``importance of the feature $i$'' in this prediction. In particular, SHAP has the property that $f(x) = \mathbb{E}[f(x)] + \sum_{i=1}^d \phi(i)$, and each $\phi(i)$ can be seen as the ``contribution'' of the feature $i$ to the total prediction $f(x)$. 

\subsection{Counterfactual Explanations}

Unlike the general idea of explanations, where the objective is to comprehend the rationale behind complex model predictions, counterfactual explanations are designed to give individuals the possibility of \textit{action}~\cite{wachter2017counterfactual}, which are also commonly called \textit{actionable recourse}~\cite{ustun2019actionable, karimi2021algorithmic}. \citet{molnar2018guide} elucidates that counterfactual explanations are an accessible method to illustrate ``what-if'' scenarios: ``What would be the prediction of the model for credit scoring, given that I increased my annual income by 10\%?''. These ``what-if'' scenarios should consider the closest world possible, as considering absurd scenarios would not contain relevant information about the model and would not be an \textit{actionable} explanation. \citet{wachter2017counterfactual} formalized the problem of generating counterfactual explanations as follows:

\begin{equation}
    \argmin_{x'} \max_\lambda \lambda (f(x') - y')^2 + c(x, x')
    \label{eq:first_ce}
\end{equation}

This formulation searches for a value $x'$ with a prediction $f(x')$ equal to the desired outcome $y'$. The function $c$ is the distance between $x$ and $x'$ and encapsulates the idea of ``closest world''. This distance function $c$ should be able to represent the ease of reaching $x'$ from $x'$ in the real world, i.e., if $x'$ is a \textbf{feasible} scenario for an individual with attributes $x$. Previous works on counterfactual explanations designed algorithms to obtain feasible counterfactual explanations~\cite{mothilal2020explaining, raimundo2022mining} and different metrics $c$ were considered~\cite{verma2024counterfactual, guidotti2024counterfactual}. Another concern is that solutions should be \textbf{plausible}, i.e., the suggested profile $x'$ should be probable in real-world data~\cite{kanamori2021dace, poyiadzi2020face}. For example, it is uncommon to observe a young individual with high investments, and a counterfactual explanation should not ask for that. Lastly, to fulfill individual preferences, it is important to provide multiple and \textbf{diverse} counterfactual explanations~\cite{mothilal2020explaining, russell2019efficient}.

Recent work has also studied how to solve the optimization problem from Eq.\ref{eq:first_ce} and similar formulations. Some of the present solutions were limited to specific models, such as linear models~\cite{ustun2019actionable, russell2019efficient}, tree ensembles~\cite{dutta2022robust, lucic2022focus, parmentier2021optimal}, image models~\cite{augustin2022diffusion} or differentiable models~\cite{mothilal2020explaining}. Model-agnostic algorithms~\cite{karimi2020model, mothilal2020explaining, raimundo2022mining, brughmans2024nice, guidotti2024stable} have been introduced, offering significant advantages due to their ability to be applied across various model types without imposing restrictions. 

\subsection{Pareto-Optimal Counterfactual Explanations}

In the classification setting, we are interested in counterfactual explanations $x' = x + a$ such that $f(x') \geq \tau$ while $f(x) < \tau$. The vector $a$ can be called an \textit{action} and has a useful interpretation: coordinate $a(i)$ represents the increase/decrease necessary on the $i$-th feature; if $a(i) = 0$, no change is necessary. Feasibility has been measured in the literature in different ways, which motivates the use of multi-objective distance functions $c : \mathbb{R}^d \times \mathbb{R}^d : \mathbb{R}^m$, where we have $m$ objectives. Considering that $\mathcal N$ is the set of indices of numerical features, one can calculate the average ``length'' of the change in the values of the features. We call this metric \textit{average continuous distance:}

\begin{equation}
    c_1(x, w) = \dfrac{1}{\# \mathcal N} \sum_{i \in \mathcal N} \dfrac{| x(i) - w(i) | }{r(i)} 
\end{equation}

Where $r(i)$ is the standard deviation of the feature $i$. Similarly, the cost of a counterfactual explanation can be represented by the maximum length of changes. We define the \textit{maximum continuous distance} as:

\begin{equation}
    c_2(x, w) = \max_{i \in \mathcal{N}} |x(i) - w(i) |
\end{equation}

When dealing with categorical features, we cannot directly measure the length of a suggested change. To consider such a variable type, an approach in the literature has been to measure the number of suggested changes, that is, the number of features with different values. This metric also arises from the idea of sparsity, which considers that moderate changes in a small number of features might be more feasible than small changes in a very large number of features. This can be calculated among all the features, not only the numerical ones. We call it as \textit{number of changes distance}:

\begin{equation}
    c_3(x, w) = \sum_{i = 1}^d \mathbb{I}_{[x(i) \neq w(i)]}
\end{equation}


\begin{algorithm}[t]
\caption{MAPOCAM}
\begin{algorithmic}[1]
\Procedure{MAPOCAM}{$f$, $x$, $\overline \A$, $c$, $k$}
    \State $Q \gets \{(\bm{0}, 0)\}$
    \While{$Q \neq \{ \}$}
        \State $a, i \gets Q.pop()$
        \If{$f(x + a) \geq \tau$}
            \State $S \gets S \cup \{x+a\}$
            \State \textbf{continue}
        \EndIf
        \If{\textit{PRUNE}($f$, $x$, $c$, $a$, $S$, $k$)}
            \textbf{continue}
        \EndIf
        \For{$v \in \overline \A(i)$}
            \State $a(i) \gets v$\;
            \State $Q \gets Q \cup \{(a, i + 1)\}$
        \EndFor
    \EndWhile
    \State \Return $S$
\EndProcedure
\Procedure{PRUNE}{$f$, $x$, $c$, $a$, $S$, $k$}
\For{$z \in S$}
    \If{$c(x, z) \preceq c(x, x+a)$} \textbf{return true} \EndIf
\EndFor
\If{$\sum_{j = 1}^{d} \mathbb{I}_{[a(j) \neq 0]} > k$} \textbf{return true} \EndIf
\State $\overline f_a \gets $ maximum prediction
\If{$\overline f_a < \tau$} \textbf{return true} \EndIf
\State \textbf{return false}
\EndProcedure
\end{algorithmic}
\label{alg:mapocam}
\end{algorithm}

When dealing with multi-objectives, the preference among two solutions $w$ and $z$ must consider all of the objectives. A solution can be seen as non-dominated if there is no other solution that is better in all objectives. These solutions can be multiple, and we can call them Pareto-Optimal solutions. 

\begin{definition}[Pareto-Optimal Solutions]
    Given a multi-objective function $c(x, \cdot) :$ $ \mathbb{R}^d \rightarrow \mathbb{R}^m$ that we want to minimize and a set of solutions $S$, a solution $w$ is Pareto-optimal if and only if $ \nexists z \in \A$ such that $c(x, z) \preceq c(x, w)$, i.e., $c_i(x, z) \leq c_i(x, w), \forall i$.
\end{definition}

The problem of generating counterfactual explanations has previously been considered by \citet{raimundo2022mining} with MAPOCAM. This algorithm uses a branch-and-bound strategy to find the Pareto-Optimal solutions in a large set $\overline \A$ of possible counterfactual explanations. $\overline \A$ is built for each sample $x$ as the product of discretized grids for each feature $\overline \A(i)$, i.e., $\overline{\A} = \overline{\A}(1) \times \dots  \times \overline{\A}(d)$. If the discretization of each feature has 10 different values, verifying which possibilities are valid counterfactual explanations would have complexity $O(10^d\delta)$ where $\delta$ is the cost of calling $f(x)$. This complexity may be infeasible even with a few features, and MAPOCAM employs different pruning strategies to reduce computational complexity.

MAPOCAM is presented in Alg.~\ref{alg:mapocam}. It searches for actions $a$ such that $f(x+a) \geq \tau$, starting with $a = \bm{0}$, and that present at most $k$ suggested changes. Then, it constructs counterfactual explanations by considering changing a feature at a time.  This procedure can be performed efficiently due to the fact that commonly defined costs for counterfactual explanations, such as the three presented above, have monotonicity with respect to $|a|$. If $|a| \preceq |\tilde a|$, we have $c(x, x+ a) \preceq c(x, x + \tilde a)$. The search strategy is designed so that if $|a| \preceq |a'|$, the action $a'$ will only be evaluated after $a$ has been evaluated. The resulting property is that the distances will not reduce as the search goes deeper. Based on that, each $\overline A(i)$ is ordered on the basis of the absolute values.

Another strategy to reduce complexity is to stop the search based on a bound of the maximum prediction $f(\cdot)$ that the current node can reach. Consider that the features $\mathcal D = \{1, \dots, j\}$ have already been evaluated in the search, and the following path will only evaluate altering the features $\{j+1, \dots, d\}$. We are interest in:
\begin{equation}
    \overline f_a = \max_{\begin{subarray}{c} \tilde a \in \overline \A \\ \tilde a(i) = a(i) , i \in \mathcal  D \end{subarray} } f(x+\tilde a)
\end{equation}

If we know that $\overline f_a < \tau$, following this path will not result in valid counterfactual explanations. To calculate this bound, MAPOCAM uses the property that some classifiers have monotonicity with respect to input $x$. If $f$ is monotonic with respect to each coordinate of $x$, the maximum value can easily be obtained by considering the action that has $a'(i) = a(i); \forall i \in \D$, and for $i \notin D$, $a'(i) = \max \overline{\A}(i)$ if $f$ is an increasing function of the $i$-th coordinate or $a'(i) = \min \overline{\A}(i)$ if it is a decreasing one. MAPOCAM is also capable of calculating the exact value of $\overline f_a$ with tree ensembles. In this scenario, the trees are traversed following the values of $x+a$, and when an split is decided by a feature not in $\mathcal D$, it should return the maximum output of following both paths.

Although MAPOCAM is capable of obtaining Pareto-optimal counterfactual explanations that are feasible, it may fail when considering plausibility. Certain solutions may not have a significant probability of being observed in the real world. Furthermore, the bound strategy depends on the fact that $f$ has monotinicity with respect to $x$, which is not valid for models such as SVM and neural networks. Our proposal, P²CE, improves the previous work by altering the algorithm so that we generate counterfactual explanations that are inlier (Sec.~\ref{sec:outlier_ce}) and supporting any model, without the monotonicity constraint (Sec.~\ref{sec:predict_max}).

\section{Generating Inlier Counterfactual Explanations}
\label{sec:outlier_ce}

Although many proposed algorithms focus on generating feasible counterfactual explanations, they may fail to ensure plausibility. This is also the case for MAPOCAM. Even if a counterfactual explanation is feasible and have a great proximity to the reference sample, it can still be outside the data distribution, as illustrated by the simple example in Fig.~\ref{fig:toy_example}. Our proposed algorithm, P²CE, was designed to generate Pareto-optimal inlier counterfactual explanations using the isolation forest~\cite{liu2008isolation} algorithm for outlier detection. 

For a dataset $X \sim P$, an outlier detection function $u(x) = 1$ if $P(X = x) \leq \varepsilon$, where $\varepsilon$ is a small positive value and $u(x) = 0$ otherwise. Isolation forest algorithm learns the function $u$ by creating multiple decision trees that are optimized to minimize the number of samples in each leaf. Samples that are ``outside'' the data distribution will be more easily isolated, and therefore the path from the root to the respective leaf will be shorter. One useful property of Isolation Forest is that, because it is a tree ensemble, it is capable of producing predictions when some coordinates of the input are absent. This is achieved by exploring both potential paths when the decision variable at a node is indeterminate. Furthermore, isolation forest has presented the best results in outlier detection tasks in recent benchmarks~\cite{bouman2024unsupervised}.

MAPOCAM searches for counterfactual explanations in a grid $\overline \A$ that is defined based on the product of discretized feature values. However, a counterfactual explanation $x+a$ may be feasible but outside of the data distribution, as $\overline \A$ is created without considering the correlation between the features. In that sense, such counterfactual explanation may not be useful to an individual, as the point $x+a$ is not reachable. Our objective is then to collect all Pareto-Optimal solutions in the set of inlier counterfactual explanations. It is important to note that this procedure will result in a different set of solutions than obtaining all Pareto-optimal actions and removing the ones from this set that are outliers. 

P²CE is presented in Alg.~\ref{alg:P²CE}. Whenever a valid solution $x+a$ is found in the search, P²CE verifies whether this counterfactual explanation is an outlier. If not, we can add this solution to the set $S$. Otherwise, this solution should not be included in the set of solutions. Although the cost of the counterfactual explanation will only increase when considering further changes in coordinate $i$, larger changes can return to being within the data distribution. This procedure could be performed with an arbitrary outlier detection $u$, such as isolation forest, autoencoder-based approaches, local outlier factor.

An additional gain in efficiency can be obtained if $u$ can evaluate whether samples $\tilde x$ with missing values are outliers. $u(\tilde x) = 1$ indicates that placing any value in the undefined coordinates will also result in an out-of-distribution sample. In that way, there is no need to consider further changes in this coordinates. This procedure is performed by P²CE on lines (20-21).

\begin{algorithm}[t]
\caption{P²CE}
\begin{algorithmic}[1]
\Procedure{P²CE}{$f$, $x$, $\overline \A$, $c$, $k$}
    \State $Q \gets \{(\bm{0}, 0)\}$
    \While{$Q \neq \{ \}$}
        \State $a, i \gets Q.pop()$
        \If{$f(x + a) > \tau$}
            \If{$u(x+a) \neq 1$}
                \State $S \gets S \cup \{x+a\}$
                \State \textbf{continue}
            \EndIf
        \EndIf
        \If{\textit{PRUNE}($f$, $u$, $x$, $c$, $a$, $i$, $S$, $k$)}
            \textbf{continue}
        \EndIf
        \For{$v \in \overline \A(i)$}
            \State $a(i) \gets v$\;
            \State $Q \gets Q \cup \{(a, i + 1)\}$
        \EndFor
    \EndWhile
    \State \Return $S$
\EndProcedure
\Procedure{PRUNE}{$f$, $u$, $x$, $c$, $a$, $i$, $S$, $k$}
\For{$z \in S$}
    \If{$c(x, z) \preceq c(x, x+a)$} \textbf{return true} \EndIf
\EndFor
\If{$\sum_{j = 1}^{d} \mathbb{I}_{[a(j) \neq 0]} > k$} \textbf{return true} \EndIf
\State $\overline f_a \gets $ maximum prediction
\If{$\overline f_a < \tau$} \textbf{return true} \EndIf
\State $\tilde x \gets [x(0) + a(0), \dots, x(i) + a(i), \textrm{NULL}, \dots, \textrm{NULL}]$
\If{$u(\tilde x) = 1$} \textbf{return true} \EndIf
\State \textbf{return false}
\EndProcedure
\end{algorithmic}
\label{alg:P²CE}
\end{algorithm}

\section{Estimating Maximum Prediction for Any Model}
\label{sec:predict_max}

As presented in Alg.~\ref{alg:P²CE}, the branch-and-bound strategy used by P²CE depends on the estimation of $\overline f_a$, the maximum prediction obtainable by further changes in action $a$. If this value is lower than $\tau$, traversing this branch becomes unnecessary. In MAPOCAM, this estimate is done under the assumption of model monotonicity, as discussed in Sec.~\ref{sec:background}. Widely used models, including KNN, neural networks, SVM, and others, do not adhere to this assumption. We present a method for estimating the maximum prediction based on SHAP values~\cite{lundberg2017unified} using the additive property of feature attributions.

In more detail, we have a partial solution $x+a$ with a set of features $\mathcal{D}$ that will not have their values altered (either they were fixed by the user or they are fixed because their values have already been altered previously in the search process). Let $\mathcal{X}_{(a, \mathcal{D})}$ denote the space where all elements have the same fixed values as $x$, i.e., if $\tilde a \in \mathcal{X}_{(a, \mathcal{D})} \implies a(i) = \tilde a(i), \forall i \in \mathcal{D}$. The goal is to determine $\overline f_a := \max_{\tilde a \in \mathcal{X}_{(a, \mathcal{D})}} f(x + \tilde a)$. If $\overline f_a < \tau$, there is no valid counterfactual in $\mathcal{X}_{(a, \mathcal{D})}$.

SHAP values have the property of being additive feature attributions, that is, $f(x) = \mathbb{E}[f(x)] + \sum_{i=1}^d \phi_i(x)$, where $\phi_i(x)$ represents the attribution of the $i$-th feature calculated for sample $x$. Notice that $\sum_{i \in \mathcal{D}} \phi_i(x)$ reflects the attribution of fixed features, while $\sum_{i \notin \mathcal{D}} \phi_i(x)$ relates to open ones. 
Consider that $\overline \phi_{p}$ is the maximum attribution of the feature $p$ for any value of $x$ in the domain of $\mathcal X$. Although we may not have access to the support of $\mathcal X$, we can approximate $\overline \phi_p$ by calculating the feature attributions of a large dataset and calculating the maximum value. Thus, we obtain a bound for $\overline f_a$:

\begin{equation}
    \begin{split}
        \max_{\tilde a \in \mathcal{X}_{(a, \mathcal{D})}}  f(x + \tilde a)  =&  \mathbb{E}[f(x)] + \sum_{i \in \mathcal{D}} \phi_i(x^\star) + \sum_{i \notin \mathcal{D}} \phi_i(x^\star)  \\
        \leq& \mathbb{E}[f(x)] + \sum_{i \in \mathcal{D}} \phi_i(x^\star) + \sum_{i \notin \mathcal{D}} \overline{\phi}_i \\
        =& \mathbb{E}[f(x)] + \sum_{i \in \mathcal{D}} \phi_i(x) + \sum_{i \notin \mathcal{D}} \overline\phi_i + \left(\sum_{i \in \mathcal{D}} \phi_i(x^\star) - \sum_{i \in \mathcal{D}} \phi_i(x) \right) \\ 
        =& \mathbb{E}[f(x)] + \sum_{i \in \mathcal{D}} \phi_i(x) + \sum_{i \notin \mathcal{D}} \overline\phi_i + R_{(x, \mathcal{D})} \\ 
        =& f(x) - \sum_{i \notin D}\phi_i(x) + \sum_{i \notin \mathcal{D}} \overline\phi_i + R_{(x, \mathcal{D})}
    \end{split}
\end{equation}

where $x^\star$ solves the maximization problem, $\phi_i(x)$ and $\phi_i(x^\star)$ are the SHAP values for the $i$-th feature for the solutions $x$ and $x^\star$ respectively. Notice that this upper bound has a residual term $R_{(x, \mathcal{D})}$ that is the difference between the attributions of fixed features (from $\mathcal{D}$) of the initial observation $x$ and the solution $x^\star$. Later, we will discuss how the residual $R(x, \mathcal{D})$ is a small value whenever $\mathcal{X}_{(a, \mathcal{D})}$ is small.

As we are using an upper bound on the optimization problem, we overstate the maximum prediction, which could increase the computational time. However, this overestimation does not compromise the quality of the solutions obtained. In Sec.~\ref{sec:experiments}, we evaluate the computing cost of our solution in real-world datasets. 

We can make a small alteration to the previously formulated problem to reduce the upper bound. This is done by considering that the generation of counterfactuals is usually done with a constraint on how many features will be changed simultaneously. Let $k$ be the number of features that can be further altered from the partial solution $x + a$ and $\mathcal{X}_{(a, \mathcal{D}, k)} = \{ \tilde a \in \mathcal{X}_{(a, \mathcal{D})} \mid \sum_{i \notin \mathcal D} \mathbb{I}[a_i \neq \tilde a_i] \leq k \}$ that is, the set of actions that have at most $k$ open features different from $a$. Let $O_k \subseteq \mathcal D^C$ be the set of $k$ open features that have the $k$-biggest $\overline \phi_i$, i.e., if $i \in O_k \implies \overline \phi_i \geq \overline \phi_j \forall j \notin O_k$. Using the same idea as from the previous proof, we can obtain a similar upper bound:

\begin{equation}
    \max_{\tilde a \in \mathcal{X}_{(a, \mathcal{D}, k)}}  f(x + \tilde a) \leq f(x) - \sum_{i \in O_k}\phi_i(x) + \sum_{i \in O_k} \overline\phi_i +  R_{(x, O_k)}
\end{equation}

As $O_k \subseteq D^C$, we have $ \left(- \sum_{i \in O_k}\phi_i(x) + \sum_{i \in O_k} \overline\phi_i \right) <  \left( - \sum_{i \in \mathcal D^C}\phi_i(x) + \sum_{i \in \mathcal D^C} \overline\phi_i \right)$ and this formulation gives an upper bound that is lower than the previous one obtained. We will estimate that $\overline f_a$ is smaller than $\tau$ more frequently, thus, reducing the number of branches that the algorithm needs to search for. 

One last detail is the residual term $\mathcal R$. This term is the difference between the feature attributions of the reference sample $x$ and the point $x^\star$ that maximizes $f(x^\star)$ ($x^\star = x + a', a' \in \mathcal{X}_{(a, \mathcal{D}, k)})$. The lower this value, the more tight our upper bound is going to be. During the search, $x$ and $x^\star$ will be close values, since only $k$ features are changed at most. Previous work by \citet{khan2024analyzing} has theoretically analyzed explainer astuteness: how much explanations can change when the input sample is altered. Their work has shown that when two points $x$ and $x'$ are sufficiently close, the SHAP attributions of these two samples will be bounded by the distance between them with high probability.

\section{Experiments}
\label{sec:experiments}

To evaluate the proposed algorithm and compare it against related techniques, we performed a set of experiments using common classifiers and real-world benchmark datasets. Our experiments evaluated the capabilities of P²CE in generating solutions that are inside data distribution and are optimal in regard to multiple objectives.

\myparagraph{Datasets} We considered three benchmark datasets of different sizes (in terms of the number of samples and features). Two datasets are from finance and are widely used to evaluate credit scoring techniques. German Credit~\cite{german} contains 1000 samples with 27 client features and a classification as good or bad risk (with 70\%-30\% distribution). Taiwan~\cite{taiwan} dataset includes information on credit card default from 30,000 clients of a Taiwanese bank with 11 features after preprocessing. We also considered Adult dataset~\cite{misc_adult_2}, a dataset with personal information and income data collected from the US 1994 census with 8 features for 48,000 individuals. To be able to support MAPOCAM, we have not included categorical features that are not binary. In each dataset, we selected a set of features that should be considered to generate counterfactual explanations, avoiding unmutable personal characteristics. In the end, German Credit had 21 considered features, Taiwan 10 and Adult had 8. 

\myparagraph{Classifiers} Our experiments included commonly employed classifier algorithms, which include a Logistic Regression (LR), Gradient-Boosting (LightGBM) and a Multi-Layer Perceptron (MLP). LR satisfies the monotonicity constraints imposed by MAPOCAM and was used in a ablation study of the inlieness of counterfactual explanations. Remaining classifiers are highly complex with non-linearities and highlight the flexibility of P²CE.

\myparagraph{Compared Algorithms} Experiments included comparisons with MAPOCAM~\cite{raimundo2022mining} to demonstrate the improvements obtained from our novel algorithm. A comparison was also made with DICE~\cite{mothilal2020explaining} and NICE~\cite{brughmans2024nice}, two popular algorithms for counterfactual explanations. DICE was initially presented as a gradient-based algorithm to generate diverse solutions, however, we employ a genetic implementation with the same objective but that is model-agnostic. It can generate an arbitrary number of counterfactual explanations and tries to maximize diversity among them. NICE is a fast approach that uses information from neighboring samples to generate counterfactual explanations. It is limited to generating only one solution for the input sample.

\myparagraph{Setting} The datasets were separated into train, validation, and test sets of size 40\%, 10\%, 50\%. Hyperparameter tuning was performed with the Optuna package~\cite{akiba2019optuna} with 50 trials, and the final models are those that achieved the highest balanced accuracy in the validation set. P²CE used an isolation forest fitted to the training data. A stronger algorithm, the extended isolation forest~\cite{hariri2019extended}, was fitted on the test data to evaluate the plausibility of counterfactual explanations. To classify outliers, we consider that all datasets have an infection rate of 5\% of outlier samples. The metrics reported in the remaining of the section are the average result of generating counterfactuals for 50 randomly selected individuals from the test set that obtained the negative prediction.

\subsection{Inlierness of Counterfactual Explanations}

The initial experiment is intended to demonstrate the ability of P²CE in generating plausible counterfactual explanations compared to MAPOCAM. To do so, we generated counterfactual explanations for the LR model, which satisfies monotonicity with respect to input $x$. In that way, P²CE does need to use SHAP to estimate $\overline f_a$, and the difference between algorithms is how they deal with outlier solutions. We also performed an ablation study by removing the procedure from lines 20-21 of Alg.~\ref{alg:P²CE} that evaluates if partial solutions can already be classified as outliers. We call this simplified version P²CE (Ablation). Both algorithms were executed considering solutions with at maximum 3 changes and considering only one objective, the average continuous distance. 

The results are depicted in Tab.~\ref{tab:experiment_lr}, including the average computation time, the average distance of the solutions, and the percentage of solutions that were outliers. Due to only using one objective, the set of Pareto-Optimal solutions is only composed of one solution. On German Credit, none of the techniques have presented outliers counterfactual explanations. This might occur because of the simplicity of the dataset, which is only composed of 1000 samples. However, in Taiwan and Adult datasets, P²CE generates fewer outlier counterfactual explanations than MAPOCAM. Although P²CE had a higher average computing time and a higher distance of solutions, this is expected. By removing solutions from $\overline \A$ (due to them being outliers), one might remove solutions that would be Pareto-Optimal. This also increases the number of solutions that must be evaluated before finding a valid one. When comparing P²CE and P²CE (Ablation), the average computing time shows that the procedure of evaluating if partial solutions are already outliers can help the algorithm in avoid unnecessary search.

\begin{table}[]
        \begin{tabular}{@{}lccc@{}}
        \toprule
                 & $c(x, x')$ & \% Outliers & Time (s)  \\ \midrule
        \multicolumn{4}{c}{German Credit}                \\ \hline
        MAPOCAM  & 0.093 ($\pm$0.081) & 0 & 15.7 ($\pm$13.6)       \\
        P²CE (Ablation) & 0.093 ($\pm$0.081) & 0 &   16.0 ($\pm$13.6)        \\ 
        P²CE & 0.093 ($\pm$0.081) & 0 & 16.2 ($\pm$14.1)        \\ \hline      
        \multicolumn{4}{c}{Taiwan}                     \\ \hline
        MAPOCAM  & 0.274 ($\pm$0.120)  & 8 & 58.8 ($\pm$51.8)  \\
        P²CE (Ablation) & 0.276 ($\pm$0.119) & 6 & 56.0 ($\pm$49.2)         \\ 
        P²CE & 0.276 ($\pm$0.119) & 6 & 51.7 ($\pm$77.1) \\ \hline
        \multicolumn{4}{c}{Adult}                     \\ \hline
        MAPOCAM  & 0.138 ($\pm$0.066) & 16 & 1.3 ($\pm$0.8) \\
        P²CE (Ablation) &  0.183 ($\pm$0.145) & 6 &  3.9 ($\pm$8.0)  \\
        P²CE & 0.171 ($\pm$0.107)  & 2 &  2.4 ($\pm$3.1)        \\ \bottomrule
        \end{tabular}
        \caption{Results of MAPOCAM and P²CE using a LR model. Reported results are the average cost and time on 50 individuals and the percentage of occurrence of outliers solutions among all individuals.}
        \label{tab:experiment_lr}
\end{table}

\subsection{One Objective in Non-linear Classifiers}

\begin{figure}
    \centering
    \includegraphics[width=\linewidth]{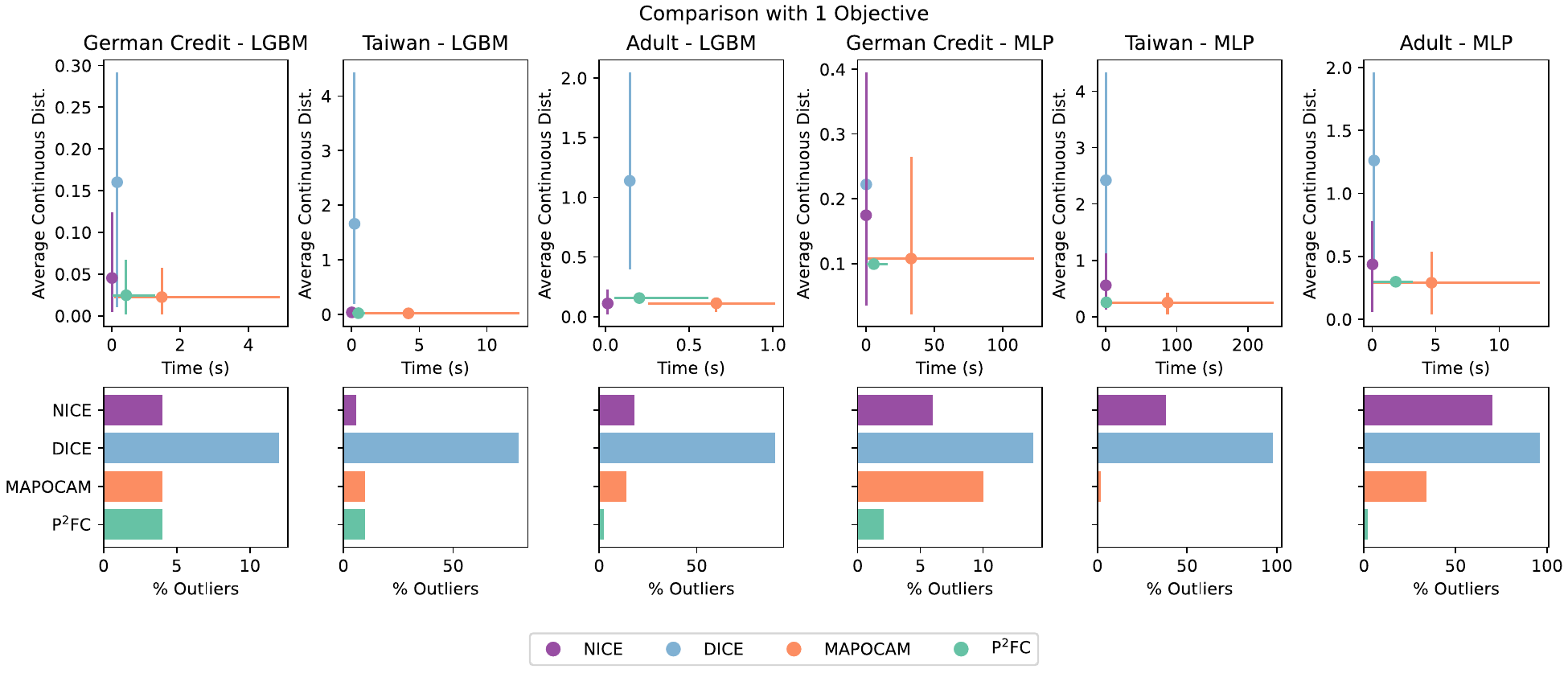}
    \caption{Resulting in computing time, distance and occurrence of outliers of P²CE and three compared algorithms with three datasets and two models. In the scatter plot, lines represent 90\% range of observed costs.}
    \label{fig:percentile_experiment}
\end{figure}

Using an LGBM and an MLP, we evaluated P²CE, MAPOCAM, DICE and NICE capabilities of generating a counterfactual with small distance (one objective). P²CE and MAPOCAM were configured with the average continuous distance objective and considering actions that change at most three features. P²CE is using the Tree Explainer for the LGBM model and Deep Explainer for the MLP model. MAPOCAM uses the adapted version for tree ensembles in the experiments with LGBM. DICE is set to generate only one solution. We present the results at Fig.~\ref{fig:percentile_experiment} separated by model and dataset. 

In all scenarios, P²CE achieved the lowest cost while maintaining highly competitive computational time. The difference in computing time between MAPOCAM and P²CE is more evident in the MLP experiments. MAPOCAM did not use an estimate of $\overline f_a$ and had to evaluate many possibilities from $\overline \A$, taking 100 seconds on average in the Taiwan dataset. P²CE was able to obtain the solutions in a few seconds. On the LGBM experiment, while MAPOCAM uses the tree structure to estimate the maximum prediction of the model, the cost of this procure is higher than the approach used by P²CE, as the average computing time shows. The results show that DICE and NICE algorithms are faster than P²CE, yet, they do not have the guarantees of generating Pareto-optimal solutions. NICE only obtained competitive costs in the Taiwan and Adult dataset with the LGBM model. Furthermore, NICE cannot handle multiple objectives or generate more than one counterfactual.

The bar plots at the bottom of Fig.~\ref{fig:percentile_experiment} show the percentage of outlier solutions generated by each algorithm. In all cases, P²CE has an outlier rate of less than  5\%, and the only other algorithm with competitive results was MAPOCAM. However, P²CE significantly exceeded the results from MAPOCAM with the MLP model.

\subsection{Multiple-objective}

To evaluate our proposal considering multiple objectives, we execute P²CE using the three objectives present in Sec.~\ref{sec:background}: average continuous distance, maximum continuous distance, and number of changes. MAPOCAM was executed using the same objectives, DICE was executed to generate 4 counterfactual explanations for each individual, and NICE was not included in this experiment, as its formulation does not consider multiple objectives. Comparing solutions with multiple objectives is not a straightforward procedure. To do so, we utilize hypervolumes, a metric defined as the ``volume'' defined between the Pareto frontier (region of Pareto-optimal solutions) and the worse performance identified. This measure is between $0$ and $1$, and the larger this volume, the better the solutions. 

Results are displayed in Tab.~\ref{tab:experiment_multi_objective}. We also included the percentage of solutions that were outliers and the average computing time. With all datasets and models, MAPOCAM presented the highest volume, closely followed by P²CE. This result was expected as MAPOCAM presents guarantees in obtaining Pareto-Optimal solutions. Even though P2CE also presents such theorical guarantees, it is not capable of obtaining MAPOCAM solutions, as some of them are outliers and also due to SHAP approximation error. However, the low distances of MAPOCAM solutions is accompanied by a high computational cost. For example, MAPOCAM required 100 seconds to generate counterfactual explanations with the Taiwan dataset and MLP, while P²CE took less than one second. In all data sets and models, DiCE represented really low hypervolume values, which is not competitive with the other two techniques.

\begin{table}[ht]
        \begin{tabular}{@{}lccc@{}}
        \toprule
         Algorithm        & Hypervolume & \% Outliers & Time (s)  \\ \midrule
        \multicolumn{4}{c}{German Credit with LGBM}                     \\ \hline
        MAPOCAM  & 0.53 ($\pm$ 0.11) & 4 &  2.23 ($\pm$3.1)      \\
        DICE & 0.31 ($\pm$ 0.21) & 20 &  0.18 ($\pm$0.0)        \\ 
        P²CE & 0.52 ($\pm$ 0.12) & 4 & 0.93 ($\pm$1.7)          \\ \hline
        \multicolumn{4}{c}{Taiwan with LGBM}                     \\ \hline
        MAPOCAM  & 0.65 ($\pm$ 0.02) & 13 &  8.0 ($\pm$12.2)      \\
        DICE & 0.41 ($\pm$ 0.15) & 83 & 1.0 ($\pm$4.0)           \\ 
        P²CE & 0.64 ($\pm$ 0.07) &  6 & 0.48 ($\pm$0.6)         \\ \hline
        \multicolumn{4}{c}{Adult with LGBM}                     \\ \hline
        MAPOCAM  & 0.58 ($\pm$ 0.06) & 14 & 0.91 ($\pm$0.5)       \\
        DICE &0.35 ($\pm$ 0.16) & 90 &  0.17 ($\pm$0.0)        \\ 
        P²CE & 0.48 ($\pm$ 0.2) & 1 &  0.94 ($\pm$1.0)        \\ \hline
        \multicolumn{4}{c}{German Credit with MLP}                     \\ \hline
        MAPOCAM  & 0.45 ($\pm$ 0.2) & 14 & 49.62 ($\pm$42.4)       \\
        DICE & 0.24 ($\pm$ 0.16) & 20 & 0.23 ($\pm$0.0)         \\ 
        P²CE & 0.44 ($\pm$ 0.21) & 5 &  9.32 ($\pm$6.0)        \\ \hline
        \multicolumn{4}{c}{Taiwan with MLP}                     \\ \hline
        MAPOCAM  & 0.41 ($\pm$ 0.16) & 11 &  118.51 ($\pm$96.8)      \\
        DICE & 0.27 ($\pm$ 0.16) & 93 &  0.28 ($\pm$0.02)        \\ 
        P²CE & 0.39 ($\pm$ 0.16) & 1 & 0.81 ($\pm$1.0)         \\ \hline
        \multicolumn{4}{c}{Adult with MLP}                     \\ \hline
        MAPOCAM  & 0.42 ($\pm$ 0.17) & 58 & 6.24 ($\pm$4.3)       \\
        DICE & 0.27 ($\pm$ 0.14) & 87 & 0.19 ($\pm$0.0)         \\ 
        P²CE & 0.4 ($\pm$ 0.18) & 11 & 2.34 ($\pm$1.4)          \\ \bottomrule
        \end{tabular}
        \caption{Average hypervolume, percentage of outliers solutions and average computing time of P²CE and two compared algorithms. The higher the hypervolume, the better. While MAPOCAM obtains higher hypervolume than MAPOFCEM, it is accompanied by significant disadvantages in generating outlier counterfactual explanations and a high computing time.}
        \label{tab:experiment_multi_objective}
\end{table}

We also present an example of the solutions identified by each of the algorithms in Tab.~\ref{tab:solutions_example}. These were the solutions generated for a specific individual from the Adult dataset that obtained the negative prediction from the MLP model. The table contains only the features that were altered by any of the solutions and also includes whether the solution is an outlier ($u(x)$) and what was the classifier prediction ($f(x)$). All solutions provided by Dice were outliers and MAPOCAM just presented one in-distribution solution. In two of the examples, Dice asks for 97 hours of weekly work, which is obsviouly an solution with no plausability. It also suggest large changes on capital gain and capital loss. Notice that while the MAPOCAM and P²CE solutions had exactly $f(x) = 0.5$, the minimum value necessary to obtain the positive prediction, the DICE solutions had higher values. P²CE can generate solutions considering other threshold values $\tau$ that can give greater confidence in the positive prediction.

\begin{table}[]
\begin{tabular}{@{}llllllll@{}}
\toprule
Capital Gain    & Capital Loss & Hours per Week & Has Degree & Is Married & Gov. Job & $u(x)$ & $f(x)$     \\ \midrule
 0 & 0& 	30 &	0 &	0 &	0  &	- & 0.32 \\ \hline
\multicolumn{7}{c}{P²CE} \\ \hline
7520&	0 &	36 &	0&	1&	0&	0&	0.50 \\
0	&0	& 52&	0&	1&	0&	0 &	0.50 \\ 
0 &	0	& 30 &	1 &	1 &	0 &	0 &	0.54 \\ \hline
\multicolumn{7}{c}{MAPOCAM}  \\ \hline
0 &	1980	&30 &	1	&0	&1	&1	& 0.50 \\
4512 &	891	&30	&0	&1	&0	&1	&0.50 \\
5264 &	792	&30&	0&	1&	0&	1&	0.50 \\
10528 &	0	&30	&0&	1&	0	&1	&0.50 \\
0	& 0&	30 &	1 &	1 &	0	& 0 &	0.54 \\ \hline
\multicolumn{7}{c}{Dice}     \\ \hline
	36470	&0	&30	&0	&1	&0	&	1	& 0.65 \\
	39877	&0	&30	&0	&0	&0	&	1	&0.53 \\
	0&	0&	97&	0& 	1 &	0 &	1 &	0.61 \\
	21794	&0	&97	&0	&0	&0		&1 &	0.63 \\\bottomrule
\end{tabular}
\caption{Example of counterfactual explanations for an random sample from the Adult dataset that received the negative prediction from the MLP model. These solutions were obtained by P²CE and MAPOCAM using multiple objectives and DICE configured to generate 4 solutions.}
\label{tab:solutions_example}
\end{table}

\section{Limitations and Future Work}

P²CE is capable of generating multiple Pareto-optimal counterfactual explanations for any model. Furthermore, the solutions are obtained with practical computing time. In this section, we discuss some of the limitations of our proposal and present future directions.

\myparagraph{SHAP Dependency} Our algorithm uses Shapley values to obtain an upper bound of the prediction of a model, as presented in Sec.~\ref{sec:predict_max}. Shapley values have exponential computing time in the number of variables of the model, and the current literature provides algorithms to approximate such values.  This is valid for SHAP algorithms such as Kernel Explainer, Permutation Explainer~\cite{lundberg2017unified} or DeepLift~\cite{lundberg2017unified, shrikumar2017learning}. Tree Explainer~\cite{lundberg2018consistent} is an exception, which leverages the tree structure to account for feature interactions. Furthermore, certain models and SHAP implementations may have a higher computing cost than others. The approximation error and computing time from SHAP algorithms should be considered as P²CE is dependent on it. The approximation error might mislead the prunning strategy, however, our solutions presented really lower costs than algorithms such as NICE and DICE, and having only higher costs than MAPOCAM, which is guaranteed to obtain Pareto-optimal solutions. Furthermore, research is still being done to obtain exact and efficient SHAP values, which we believe will further improve the capabilities of P²CE.

\myparagraph{Discretized Grid} P²CE searches for counterfactual explanations within a discrete grid of possibilities, and the granularity of this grid limits how close a solution can be identified. Although increasing granularity will result in better optimal solutions, it will also increase search duration. Due to the practical use of counterfactual explanations, a really thin granularity is not necessary. For example, on a credit loan application, a counterfactual explanation that suggests an increase in income by $1000$ is equally as good as one that suggests an increase by $998$. P²CE allows the grid to be set with great control for each application by selecting minimum and maximum values and the specific granularity of each future. A possible future direction is developing an adaptive grid, which will automatically increase the granularity of features that can indeed alter the predictions and decreasing the granularity of features that have a low impact on the outcome. This procedure could also take advantage of SHAP values to do so.

\section{Conclusion}

In this work, we presented P²CE, a novel algorithm for generating Pareto-optimal counterfactual explanations. P²CE distinguishes itself through its model-agnostic design and distribution awareness, achieved by integrating an isolation forest for outlier detection and leveraging SHAP values for efficient computation. This approach allows P²CE to provide users with a diverse set of actionable and plausible explanations, even for complex nonlinear models. Our empirical evaluation across various datasets and scenarios demonstrated that P²CE achieves Pareto-optimal solutions with superior performance in terms of computational cost and outlier detection, enabling faster and more informed decision-making in real-world applications.


\bibliography{paper}

\end{document}